%% file: emnlp2022.tex
\pdfoutput=1

\documentclass[11pt]{article}

\usepackage[]{EMNLP2022}

\usepackage{times}
\usepackage{latexsym}

\usepackage{graphicx}
\usepackage{soul}
\usepackage{url}
\usepackage{hyperref}
\usepackage{amsmath}
\usepackage{amsthm}
\usepackage{booktabs}
\usepackage{algorithm}
\usepackage{algorithmic}
\usepackage{array}
\usepackage{color, colortbl}
\usepackage{multirow}
\usepackage{setspace}
\usepackage{autobreak}
\usepackage{amsfonts}
\usepackage{amssymb}
\usepackage{amsthm,amsmath}
\usepackage{mathrsfs}
\usepackage{multirow}
\usepackage{subfigure}
\usepackage{makecell}
\usepackage{pifont}
\usepackage{fontawesome}
\usepackage{bm}

\usepackage[T1]{fontenc}

\usepackage[utf8]{inputenc}

\usepackage{microtype}

\usepackage{inconsolata}

\title{Kernel-Whitening: Overcome Dataset Bias with Isotropic Sentence Embedding}

\author{
    { Songyang Gao$^{1}$, \ \ Shihan Dou$^{1}$, \ \ Qi Zhang$^{12}$\thanks{{ }{ }{ }Corresponding author.} , \ \ Xuanjing Huang$^{12}$} \\
    \normalsize{$^1$  School of Computer Science, Fudan University, Shanghai, China} \\
    \normalsize{$^2$  Shanghai Key Laboratory of Intelligent Information Processing, Shanghai, China} \\
    \normalsize{ \{gaosy21, shdou21\}@m.fudan.edu.cn}
}

\begin{document}
\maketitle

\input{outline/abstract}

\input{outline/introdiction}

\input{outline/motivation}
\input{outline/method}

\input{outline/experiments}

\input{outline/Discussion}

\input{outline/related}
\input{outline/conclusion}
\input{outline/limitation}

\section*{Acknowledgements}
The authors wish to thank the anonymous reviewers for their helpful comments. This work was partially funded by National Natural Science Foundation of China (No. 62076069, 61976056, 61906176), Program of Shanghai Academic Research Leader (No. 22XD1401100), and Beijing Academy of Artificial Intelligence (BAAI).

\bibliography{anthology,custom}
\bibliographystyle{acl_natbib}

\end{document}

%% file: outline/abstract.tex
\begin{abstract}
Dataset bias has attracted increasing attention recently for its detrimental effect on the generalization ability of fine-tuned models. The current mainstream solution is designing an additional shallow model to pre-identify biased instances. However, such two-stage methods scale up the computational complexity of training process and obstruct valid feature information while mitigating bias.
To address this issue, we utilize the representation normalization method which aims at disentangling the correlations between features of encoded sentences. We find it also promising in eliminating the bias problem by providing isotropic data distribution. We further propose Kernel-Whitening, a $\rm Nystr\ddot om$ kernel approximation method to achieve more thorough debiasing on nonlinear spurious correlations. Our framework is end-to-end with similar time consumption to fine-tuning. Experiments show that 
Kernel-Whitening significantly improves the performance of BERT on out-of-distribution datasets while maintaining in-distribution accuracy.

\end{abstract}

%% file: outline/introdiction.tex
\section{Introduction}
Despite remarkable performance on NLP tasks, pre-trained language models, like BERT, suffer sharp performance degradation in out-of-distribution (OOD) settings \cite{mccoy2019right}. The above defect roots in the excessive reliance on spurious correlations, which is widely found in crowdsourcing-built datasets \cite{gururangan2018annotation}. These phenomena are donated as dataset bias problem \cite{he2019unlearn}.
A line of works attempts to tackle this problem by down-weighting bias training examples to discourage the main model from adopting recognized biases, including example reweighting \cite{schuster2019towards}, confidence regularization \cite{utama2020mind}, or model ensembling \cite{clark2019don}.

However, aforementioned methods over-depend on researchers' intuition and task-specific insights to characterize spurious correlations, causing unrecognized bias patterns to remain in individual dataset \cite{sharma2018tackling}. Such assumption that dataset biases are known as a prior has been relaxed by limited capacity models \cite{utama2020towards} or early training  \cite{tu2020empirical} in recent works. These approaches still rely on extra shallow models, which are not end-to-end, and weak-weighted bias samples simultaneously obstruct learning from their non-bias parts \cite{wen2021debiased}.

\begin{figure}[t]
\centerline{\includegraphics[width=0.485\textwidth]{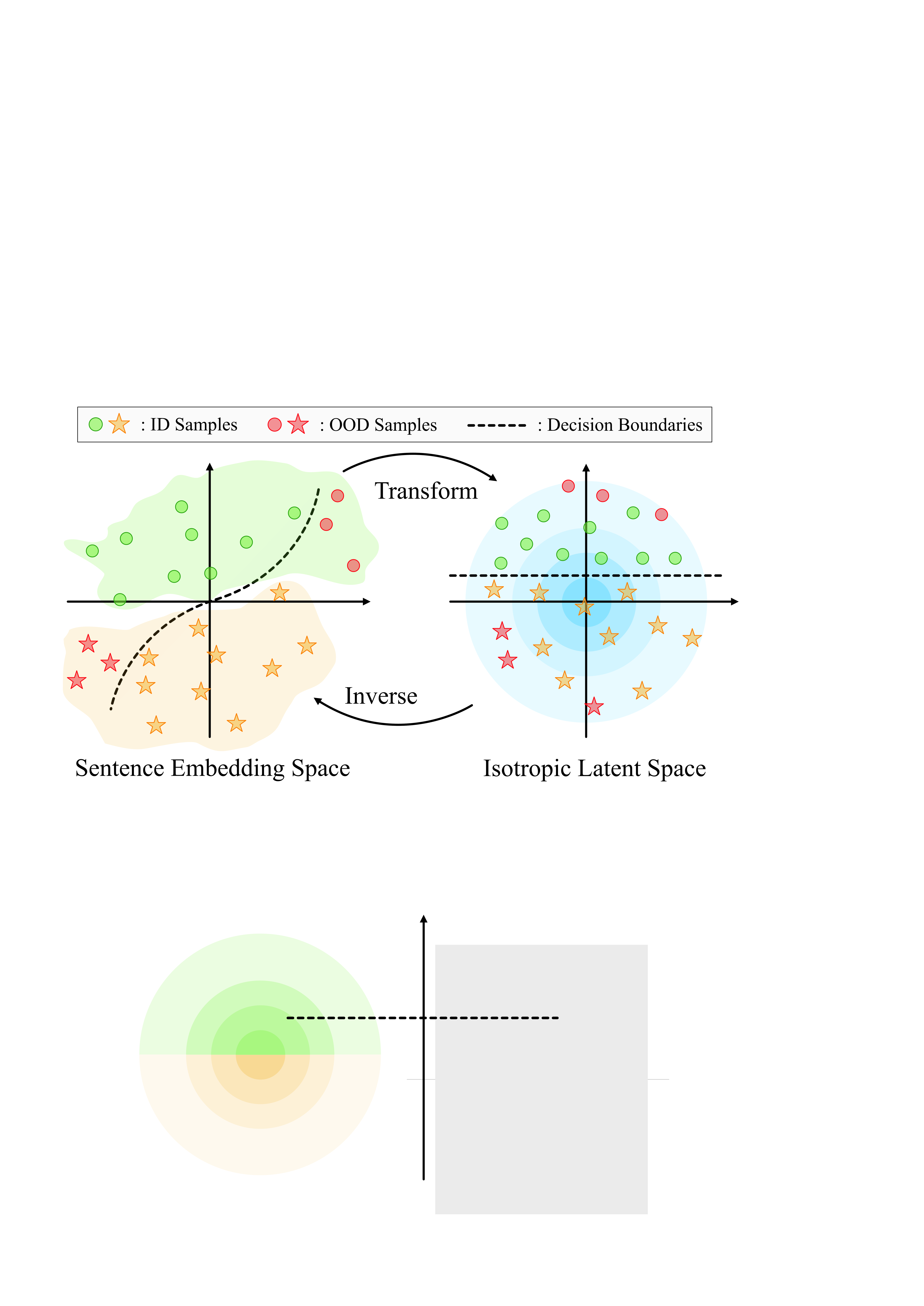}}
\caption{Illustration of Kernel-Whitening. The vertical and horizontal axes represent the valid and invalid features, respectively.
Uneven sample distribution induces a bias decision boundary, resulting in errors on out-of-distribution data. The normalization method maps the data to isotropic latent space, where the new boundary is uncorrelated to redundant features, providing better generalization.}
\label{fig:2}
\vspace{-0.4em}
\end{figure}

Instead of designing an extra model as previous attempts did, in this work, we propose a novel end-to-end framework, Kernel-Whitening, to significantly improve OOD performance while maintaining a similar computational cost as conventional BERT fine-tuning. The BERT-whitening \cite{su2021whitening} and BERT-flow \cite{li2020sentence} methods are effective normalization techniques to obtain better semantic representation. BERT-whitening calculates a linear operator\footnote{A positive definite symmetric matrix.} with SVD decomposition \cite{golub1970handbook} to transform the sentence representation to follow a distribution with respect to the standard normal distribution. 
BERT-flow introduces normalized flow \cite{rezende2015variational} to perform similar transformations. Particularly, we find that the normalization method is also promising in improving the generalization ability of fine-tuned models by eliminating spurious correlations in training datasets. Despite the significant improvement over the OOD datasets, the linear transformation of BERT-whitening is not capable of dealing with nonlinear dependencies between features. 
Meanwhile, flow-based methods require a complex inference process, which scales up the training costs.

In an attempt to eliminate nonlinear correlations while maintaining low training expenditure, we introduce kernel methods to naturally reconstruct a set of sentence representations only with linear correlation \cite{achlioptas2001sampling}. However, traditional kernel methods focus only on data similarities without providing explicit mapping operators, therefore, we use the $\rm Nystr\ddot om \  $ approximation \cite{xu2015nystrom} to obtain low-rank kernel estimations. In general, we transform training data to an isotropic Gaussian distribution without affecting topological relationships between data points.

Kernel-Whitening\footnote{Our code is available at \url{https://github.com/SleepThroughDifficulties/KernelWhitening}.} achieves competitive performance on generalization tasks. Experiments on eight datasets demonstrate that our method can improve the accuracy by 7\%-11\% on OOD datasets. In addition, the analysis of sentence representation proves that our method effectively removes the spurious correlations between dimensional features, which are known to be the direct cause of the dataset bias problem. Overall, our main contributions are as follows:

\begin{itemize}
\item We propose a novel framework, Kernel-Whitening, which ameliorates the bias problem by transforming sentence representation into isotropic distribution with similar time to fine-tuning.
\item We introduce a kernel estimation algorithm, $i.e.$, $\rm Nystr\ddot om $ approximation, to alleviate normalization methods from the trade between complex arithmetic and disentangle effects.
\item We conduct comprehensive experiments on debiasing tasks to verify the effectiveness of normalization methods for overcoming spurious correlations. 

\end{itemize}

%% file: outline/motivation.tex
\section{How Do Normalization Methods Provide Better Generalization}
In this section, we discuss the negative impacts of dataset bias on the model's generalization ability, and subsequently, how normalization methods lead to better performance in OOD settings.
\label{sec:2}
\subsection{Illustrate Dataset Bias from Feature Perspective}
\begin{figure}[t]
\centerline{\includegraphics[width=0.42\textwidth]{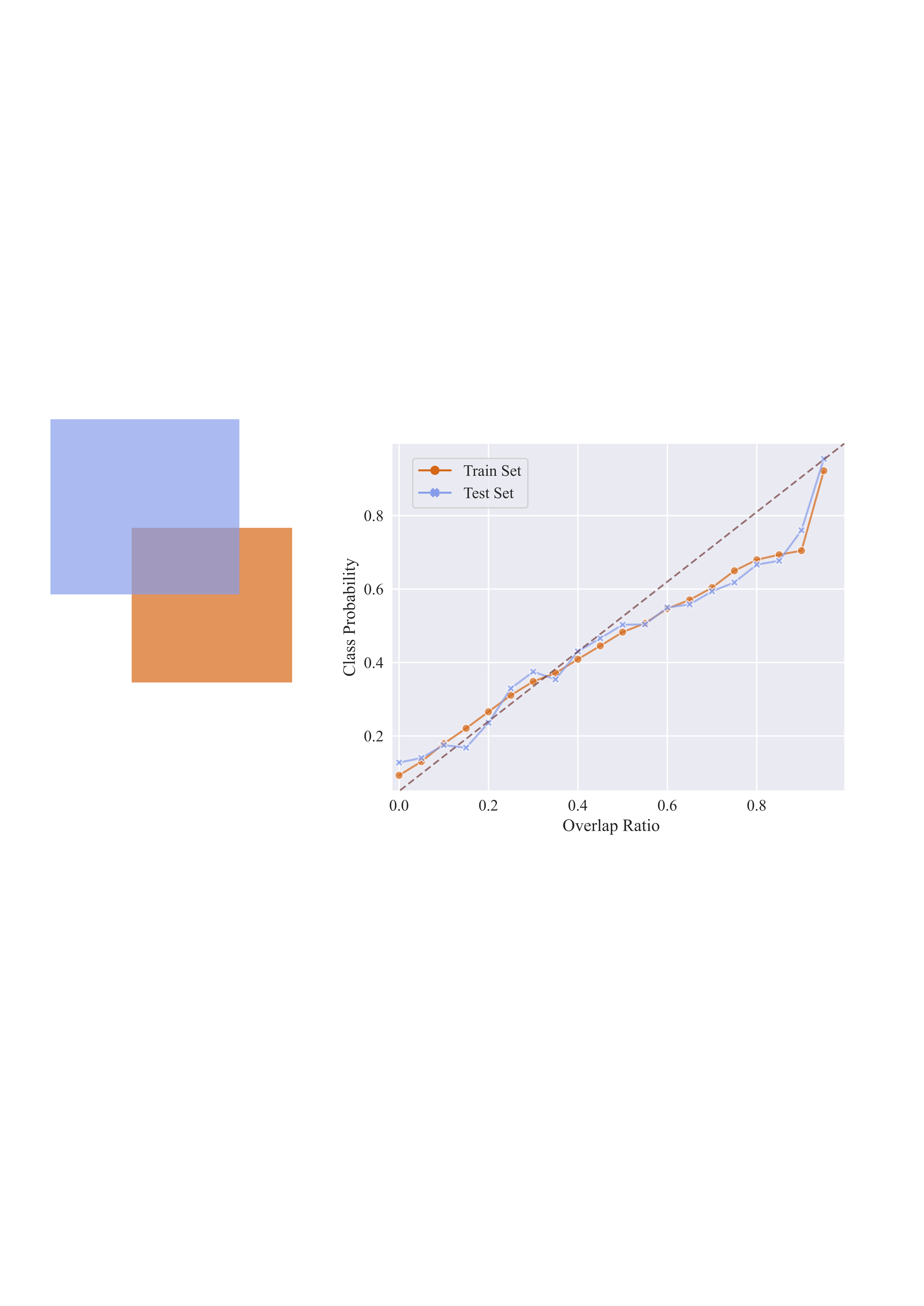}}
\caption{ Lable probability with increasing lexical overlap for entailment label on MNLI dataset. }
\label{fig:1}
\vspace{-1em}
\end{figure}

We first interpret dataset bias as triggered by the imbalance distribution of training data in feature space. Figure \ref{fig:1} shows an empirical analysis on MNLI dataset \cite{williams2018broad}. Both train and test sets exhibit a strong positive correlation between the word overlap ratio and the class portion of the entailment label, which enable models to achieve high accuracy scores without modelling semantic information. \citet{mccoy2019right} proposed HANS, a label-fair test set ($i.e.$, labels are proportionally consistent with different degrees of word overlap) to investigate the generalizability of models fine-tuned with above-mentioned bias, 
the model provides over 20\% accuracy degradation when the specific literal heuristic could not be utilized for prediction.

Formally, given input data $(X,Y)$ and bias dataset $D$, the training process can be formulated as follows:
\begin{align}
    P(Y |X^L,X^P)&=\frac{P(Y |X^L) P(X^P|X^L,Y) } {P(X^P)} \nonumber\\
    &= P(Y |X^L),
\end{align}
where $X^L$ represents the task-related features, and $X^P$ represents other irrelevant features.
That is, the model learns $X^L$ from the label pair $(X,Y)$. By modeling the conditional distribution of labels relative to the input, the model extracts valid features for specific tasks.

However, previous works argue that construction process of bias dataset introduces spurious correlations between $(X^L|D)$ and $(X^P|D)$ \cite{gururangan2018annotation}: 
\begin{align}\label{formula:2}
    P(X^P |X^L,D)&=\frac{P(X^P,X^L|D) }{P(X^L|D) } \nonumber \\
    & > P(X^P |D).
\end{align}

Therefore, the actual training objective on dataset $D$ is the posterior between feature distribution $(X^L,X^P|D)$ and label $Y$, $i.e.$, $P(Y|X^L,X^P;D)$. By Eq.\ref{formula:2}, 
irrelevant features increase the confidence for specific labels:

\begin{small}
\begin{align}\label{formula:4}
    P(Y |X^L,X^P,D)&=\frac{P(Y |X^L,D) P(X^P|X^L,D) } {P(X^P)}\nonumber \\
    &> P(Y |X^L,D) \nonumber\\
    &= P(Y |X^L).
\end{align}
\end{small}

Such overconfidence does not perturb the model effect on the test set, which has a similar distribution to the training set. However, out-of-distribution data follow the correct distribution $P(Y |X^L)$, and are therefore classified as the bias label ($e.g.$, "entailment" in MNLI dataset), even if they have different relations.

\subsection{Isotropic Representation Leads to Better Generalization}

According to the definition described in Eq.\ref{formula:4}, dataset bias causes deep networks to fit the dataset-specific distribution, which impairs the generalization performance.
Normalization methods intervene in the above problem by reconstructing the feature space. When sentences are encoded by pre-train model, the embedding representations are transformed into isotropic distribution, $e.g.$, standard normal distribution. Suppose data $x$ and prior $u$ satisfy:
\begin{align}
    x = f_{\theta}(u),\  u\sim P_{\mathcal{U}}(u),
\end{align}
where $\mathcal{U}$ represents the latent space of isotropic $u$, and $f$ is an invertible function. The probabilistic density function of original data on the transformed space can be calculated as follows:
\begin{align}\label{formula:5}
    P_{\mathcal{X}}(\mathbf{x})=\left|\operatorname{det} \frac{\partial f_{\theta}^{-1}(\mathbf{x})}{\partial \mathbf{x}}\right| P_{\mathcal{U}}\left(f_{\theta}^{-1}(\mathbf{x})\right).
\end{align}

By Eq.\ref{formula:5}, the distribution of training data is transformed into isotropic ones. Subsequently, suppose $X^L_u$ and $X^P_u$ are the latent representations of $X^L$ and $X^P$, the spurious correlations between valid features and invalid features are eliminated:

\begin{small}
\begin{align}

    P_{\mathcal{U}}(Y|X^P_u,X^L_u,D) &= \frac{P_{\mathcal{U}}(Y|X^L_u,D)P_{\mathcal{U}}(X^L_u,X^P_u|D)}{P_{\mathcal{U}}(X^P_u)P_{\mathcal{U}}(X^L_u|D)} \nonumber \\
    &= P_{\mathcal{U}}(Y|X^L_u),
\end{align}
\end{small}

With isotropic data distribution, a decision boundary independent of the redundant features is obtained, which provides better generalization on the OOD samples.

\subsection{Minor Weakness for BERT-flow and BERT-whitening}
The BERT-flow method learns a flow-based generative model to fit the transform function $ f_{\theta}$, and BERT-whitening method computes the inverse linear operator with SVD decomposition of the covariance matrix. Despite the decent effect of representing normalization methods, BERT-flow requires multiple convolutional layers to find the appropriate transformation function, which increases the difficulty and time consumption of the training process. When fine-tuned on a tiny dataset, the flow layer encountered obstacles in providing reasonable transform results. Moreover, the BERT-whitening method focuses only on eliminating linear correlations between features, which is ineffective in alleviating the nonlinear correlation problem. In an attempt to ease the effort of training and provide faster but thorough transformation, we propose a novel normalization framework by kernel approximation, which is detail discussed in the next section.

%% file: outline/method.tex
\section{Distribution Generalization with Kernel Approximation}
In this section, we show our end-to-end framework named Kernel-Whitening. We first introduce $\rm {Nystr\ddot om}$ kernel estimation algorithm, and subsequently how to apply such approximation method on debiasing tasks.

\subsection{ $ \mathbf{Nystr\ddot om}$ Kernel Estimation} \label{sec:3.1}

We first elaborate kernel trick, which constructs new linearly differentiable properties by mapping the original feature space onto a high-dimensional RKHS \cite{alvarez2012kernels}. Given a set of training data $X = \left\{x_{i} \in \mathbb{R}^{d}, i=1, \ldots, n\right\}$, the kernel method maps $X$ onto a dot product space $\mathcal{H}$ using  $\phi:X\xrightarrow{}\mathcal{H}$. Generally, the dimension of $\mathcal{H}$ can be so large that the mapping function cannot be obtained explicitly. Nevertheless, the dot product result can be represented by a positive definite kernel $k$, $i.e.$, a function $k$ satisfies:
\begin{align}
    k(x_k,x_j) &= \left< \phi(x_k),\phi(x_j)\right> \nonumber \\
    &= \sum_{i=1}^{N} \lambda_{i} \phi_{i}(x_k) \phi_{i}(x_j),
\end{align}
where $\lambda_i$ and $\phi_{i}$ denotes the eigenvalues and eigenfunctions of kernel operator $k$, and $N$ denotes their number. With finite dataset $\left\{x_{i} \in \mathbb{R}^{d}, i=1, \ldots, n\right\}$, such decomposition can be replaced with empirical estimation as follows:
\begin{align} \label{formula:8}
    \frac{1}{n} \sum_{j=1}^{n} k\left(\boldsymbol{x_k}, \boldsymbol{x}_{j}\right) \phi_{i}\left(\boldsymbol{x}_{j}\right) = \lambda_{i} \phi_{i}(\boldsymbol{x_k}).
\end{align}

Eq.\ref{formula:8} indicates a spectral decomposition of kernel matrix $G$, which satisfy $G_{k,j} = k(x_k,x_j)$. Considering the SVD decomposition of $G$ as
\begin{align}\label{formula:9}
    G = W\Sigma W^T,
\end{align}
where $W$ is an orthogonal matrix, and $\Sigma$ is a diagonal matrix with positive diagonal elements. When matching Eq.\ref{formula:8} with Eq.\ref{formula:9}, the mapping operator $\phi$ is therefore denoted as:
\begin{align}\label{formula:10}
    \phi(x_j) =  G_{j,\  :}W\Sigma^{\frac{1}{2}}.
\end{align}

However, existing datasets often contain thousands to hundreds of thousands samples, which makes it impossible to directly calculate the SVD decomposition. Therefore, we introduce $ \rm{Nystr\ddot om}$ method \cite{williams2001using} to provide a low-rank estimation of kernel matrix. 

Suppose $S$ to be a sampled subset of $X$, the kernel matrix $G$ can be represented as:
\begin{align}
    G = \left [ \begin{array}{cc}
        G_s\  G_r^T    \\
        G_r\  G_{x}
    \end{array} \right ],
\end{align}
where $G_s$ denotes the gram matrix of subset $S$. The $W$ and $\Sigma$ can subsequently approximated by:
\begin{align}
    W \approx \left [ \begin{array}{cc}
        G_s \  G_r  
    \end{array} \right ]^T,\  \Sigma \approx G_s.
\end{align}
The reconstructed $ \rm{Nystr\ddot om}$ representation of single example $\textbf{x}$ is as follows:
\begin{align} \label{formula:13}
    \phi(\mathbf{x}) =  G_{\textbf{x},S}G_s^{- \frac{1}{2}},
\end{align}
where $G_{\textbf{x},S} = [k(x_1,\textbf{x}),...,k(x_s,\textbf{x})]^T$ for $x_i \in S$. By estimating high-dimensional representations of the training samples we obtain a linearly divisible distribution, which can be normalized with a linear transformation. 

\subsection{Batch Iterate for Global Approximation}\label{Sec:3.2}
In Section \ref{sec:3.1}, we elaborate the standard $ \rm{Nystr\ddot om}$ approach to processing input data. The difference is, the subset $S$ in the traditional approach, $i.e.$, kernel SVM, usually contains hundreds of elements, while deep networks are trained on smaller batches ($e.g.$, 32 for Kernel-Whitening) with stochastic gradient descent (SGD) optimizer. 
The insufficient samples compromise the information of reconstructed representations, making the improvement inconspicuous when directly applying $ \rm{Nystr\ddot om}$ methods to debiasing tasks.

In attempt to introduce global information while processing batch data, we design preservation and reloading structures to extend the dimension of low-rank kernel matrix. For each batch, we calculate $ \rm{Nystr\ddot om}$ matrix with batch features $Z_L^t$ and global features $Z_f^t$, which represent the principal components of data distribution in step $t$. Giving input data $Z_L^t$ contains $L$ instances, the extended representation is gived with Eq.\ref{formula:13}:
\begin{align}\label{formula:14}
    \phi(Z_L^t) =  G_{0:L,:}G^{- \frac{1}{2}},
\end{align}
where $G$ denotes the kernel matrix generated by $\left[Z_L^t,Z_f^t   \right]^T$, and $G_{0:L,:}$ denotes the first $L$ rows of $G$. Especially, we select Radial Basis Function (RBF) kernel in Kernel-Whitening method.

Noticing that the kernel estimation method only projects the data points into a linearly separable space, we further normalize the distribution with a linear transformation.
The reconstructed representation $\phi(Z_L^t)$ is subsequently weighted under the supervised signal of Hilbert-Schmidt independence criterion (HSIC) \cite{wang2021learning}, which is an adequate indicator for estimating the mutual independence between features. The optimal weight $ W^{*}$ is calculated by:
\begin{align}\label{formula:15}
    W^{*} = \underset{W \in \mathcal{R}^L}{\arg\min}  \left\|\tilde{\Sigma}_{G,W}\right\|_{F}^{2},  
\end{align}    
where $W = \{W_i\}^n_{i=1}$ denotes the sample weight vector, and $\tilde{\Sigma}_{G,W}$ represents the empirical estimation for covariance between features:
\begin{align}
\tilde{\Sigma}_{G,W}&=\frac{1}{L-1} \sum_{1<i<j<L}
 F_i   ^{T}  
 F_j \\
 F_i &= WG_{:,i} - \frac{1}{n}\sum_{j=1}^n W_jG_{j,:}.
\end{align}


At the end of each iteration, we update the global features with local information $Z_L^t$ to catch reasonable basis vectors:
\begin{align}\label{formula:18}
    Z_{f}^{t+1} = \alpha_i Z_{f}^t + (1-\alpha_i)Z_L^t,
\end{align}
where $\alpha_i$ denotes the attenuated factor controlling the importance of local information. 

\subsection{Training Objective}
In Section \ref{Sec:3.2}, we show how to obtain the reconstructed feature representation $\phi(Z_L^t)$  and weighting parameters $W^{*}$. Subsequently, we use above results to train on original BERT models. The final train loss of Kernel-Whitening is denoted as:

\begin{align}\label{formula:19}
    Loss = \sum_{i=1}^L   W^{*}_i f(\phi(Z_L^t)_i, y_i),
\end{align}
where $f( \cdot , \cdot)$ represents the cross-entropy loss with input $\phi(Z_L^t)_i$ and it's corresponding label $y_i$. Our detailed algorithm implementation is shown in Algorithm \ref{alg:Framwork}.

\renewcommand{\algorithmicrequire}{ \textbf{Input:}} 
\renewcommand{\algorithmicensure}{ \textbf{Output:}} 
\begin{algorithm}[ht]
\caption{ Framework of Kernel-Whitening for our system.}
\label{alg:Framwork}
\begin{algorithmic}[1] 
\REQUIRE ~~ 
The set of pooler output for current batch, $Z = \{Z_i\}^n_{i=1}$;
The set of global features, $Z_f^t$;
The classifier for specific task, $f( \cdot , \cdot)$.
\STATE {Compute kernel matrix $G$ by $[Z,Z_f^t]^T$}
\STATE {Compute $W, \Sigma$ by SVD(G)}
\STATE {Compute representation $\phi(Z)$ by $G, W, \Sigma$ with Eq.\ref{formula:10}, \ref{formula:14}}
\STATE {Compute weight $W^{*}$ by HSIC(G,W) with Eq.\ref{formula:15}}
\STATE {Update global features $Z_f^t$ with Z with Eq.\ref{formula:18}}
\STATE {Compute training loss $L$ by classifier $f( \cdot , \cdot)$, $W^{*}$ and $\phi(Z)$ with Eq.\ref{formula:19}}
\ENSURE ~~\\
Updated global features, $Z_f^{t+1}$; \\
Training loss, $L$\\ 
\end{algorithmic}
\end{algorithm}

%% file: outline/experiments.tex
\section{Experiments} 
In this section, we provide a comprehensive analysis of Kernel-Whitening and the other two normalization methods ($i.e.$, BERT-flow \cite{li2020sentence} \footnote{The code of BERT-flow is available at \url{https://github.com/ bohanli/BERT-flow}.} and BERT-whitening \cite{su2021whitening}) through extensive experiments on three tasks.

\begin{table*}[t]
  \centering
  \begin{spacing}{1.03}
    \setlength{\tabcolsep}{1.4mm}{
\begin{tabular}{l|cc|ccc|ccc}
\toprule
\multicolumn{1}{c|}{\multirow{2}[4]{*}{\textbf{Model}}} & \multicolumn{2}{c|}{\textbf{MNLI}} & \multicolumn{3}{c|}{\textbf{FEVER}} & \multicolumn{3}{c}{\textbf{QQP}} \\
\cmidrule{2-9}      & ID    & HANS  & ID    & Symm. v1 & Symm. v2 & ID    & PAWS dupl  & $\neg$ dupl \\
\midrule
BERT-base & 84.3  & 61.1  & 85.4  & 55.2  & 63.1  & 91    & 96.9  & 9.8 \\
\midrule
Reweighting & 83.5  & 69.2  & 84.6  & 61.7  & 66.5  & 85.5  & 49.7  & 51.2 \\
Product-of-Experts & 84.1  & 66.3  & 82.3  & 62.0    & 65.9  & 88.8  & 50.3  & 61.2 \\
$\rm {{\mbox{PoE}}_{\mbox{\textbf{cross-entropy}}}}$ & 83.6  & 67.3  & 85.7  & 57.7  & 61.4  & -     & -     & - \\
$\rm {{\mbox{PoE}}_{\mbox{\textbf{self-debias}}}}$ & 80.7  & 68.5  & 85.4  & 59.7  & 65.3  & 77.4  & 44.1  & \textbf{69.4} \\
Learned-Mixin & 84.2  & 64.0    & 83.3  & 60.4  & 64.9  & 86.6  & 69.7  & 51.7 \\
Conf-reg & 84.3  & 69.1  & 86.4  & 60.5  & 66.2  & 89.1  & 91.0    & 19.8 \\
$\rm {{\mbox{Conf-reg}}_{\mbox{\textbf{self-debias}}}}$ & 84.3  & 67.1  & 87.6  & 59.8    & 66.0  & 85.0  & 48.8  & 28.7 \\
MoCaD  & 82.3  & 70.7  & 87.1  & 65.9  & 69.1  & -     & -     & - \\
\midrule
\rowcolor[RGB]{226,240,233}\textbf{BERT-flow*}   & 82.5     & 69.1     & 85.7     & 64.4     & 67.2     & -     & -    & - \\
\rowcolor[RGB]{226,240,233}\textbf{BERT-whitening-64*}   & 84.3     & 69.9     & 86.7     & 65.7     & 69.5     & 87.5     & 82.5     & 34.7 \\
\rowcolor[RGB]{226,240,233}\textbf{BERT-whitening-384*}    & 83.7     & 70.4     & 85.9     & 65.9     & 68.3     & 87.3     & 75.3     & 41.3 \\
\midrule
\rowcolor[RGB]{226,240,233}\textbf{Kernel-Whitening-64*}    & \textbf{84.4}     & 70.1     & 87.4     & 65.1     & 69.8     & \textbf{90.8}     & \textbf{91.2}     & 32.4 \\
\rowcolor[RGB]{226,240,233}\textbf{Kernel-Whitening-384*}   & 83.9     & \textbf{70.9}     & \textbf{87.8}     & \textbf{66.2}     & \textbf{70.3}     & 87.6     & 72.7     & 43.2 \\
\bottomrule
\end{tabular}%
\caption{Models evaluation on MNLI, FEVER, QQP, and their respective challenge test sets. The performance of the three normalized models is shown in cyan, the model name with asterisks represents the experimental results on our machine. The best results on each dataset are bolded.}\label{tab:exp1}
     }%
    \end{spacing}
  \vspace{-1em}
\end{table*}%

\subsection{Baseline Methods}
Our method is compared with previous works as follows:
\begin{itemize}
\item{\citet{clark2019don}} (Reweighting and Learned-Mixin), which predicts confidence for each sample and down-weights problematic data.
\item{\citet{sanh2020learning}} (Product-of-Experts and $\rm {{\mbox{PoE}}_{\mbox{\textbf{cross-entropy}}}}$), which trains limited capacity models as experts to debias without explicitly identifying dataset bias.
\item{\citet{utama2020towards}} ($\rm {{\mbox{PoE}}_{\mbox{\textbf{self-debias}}}}$ and $\rm {{\mbox{Conf-reg}}_{\mbox{\textbf{self-debias}}}}$), which uses a shallow model to identify biased samples and focus the main model on them.
\item{\citet{utama2020mind}}  (Conf-reg), which uses confidence regularization to discourage models from exploiting biases.
\item{\citet{xiong2021uncertainty}} (MoCaD), which produces uncertainty estimations to achieve a three-stage ensemble-based debiasing framework.

\end{itemize}







\begin{table}[t] 
\small
  \centering
  \begin{spacing}{1.2}
    \setlength{\tabcolsep}{1.7mm}{
    \begin{tabular}{l|c|c}
    \toprule
    \multicolumn{1}{c|}{\textbf{Model}} & \textbf{\makecell{Requires Prior \\ Knowledge}} & \textbf{Extra Model} \\
    \midrule
    Reweighting & \color[RGB]{200,85,80}\ding{52}     & \color[RGB]{200,85,80}\ding{52} \\
    Product-of-Experts & \color[RGB]{200,85,80}\ding{52}     & \color[RGB]{200,85,80}\ding{52} \\
    $\rm {{\mbox{PoE}}_{\mbox{\textbf{cross-entropy}}}}$ & \color[RGB]{200,85,80}\ding{52}     & \color[RGB]{200,85,80}\ding{52} \\
    $\rm {{\mbox{PoE}}_{\mbox{\textbf{self-debias}}}}$ & \color[RGB]{40,160,70}\faTimes     & \color[RGB]{200,85,80}\ding{52} \\
    Learned-Mixin & \color[RGB]{200,85,80}\ding{52}     & \color[RGB]{200,85,80}\ding{52} \\
    conf-reg & \color[RGB]{200,85,80}\ding{52}     & \color[RGB]{200,85,80}\ding{52} \\
    $\rm {{\mbox{Conf-reg}}_{\mbox{\textbf{self-debias}}}}$ & \color[RGB]{40,160,70}\faTimes     & \color[RGB]{200,85,80}\ding{52} \\
    MoCaD & \color[RGB]{200,85,80}\ding{52}     & \color[RGB]{200,85,80}\ding{52} \\
    \midrule
    \textbf{Kernel-Whitening} & \color[RGB]{40,160,70}\faTimes     & \color[RGB]{40,160,70}\faTimes \\
    \bottomrule
    \end{tabular}%

     }
    \end{spacing}
  \caption{ Details of the state-of-the-art debiasing methods used to compare with Kernel-Whitening. Our model is end-to-end while not requiring prior knowledge of biases or additional shallow models.}\label{tab:detail}%
  \vspace{-1em}
\end{table}%


\subsection{Datasets and Metrics}

We conduct experiments on three tasks: natural language inference, fact verification, and paraphrase identification. Each task contains in-distribution and out-of-distribution datasets. 

For the NLI task,
we conduct experiments on the Multi-Genre Natural Language Inference (MNLI) dataset \cite{nangia2017repeval} and HANS \cite{mccoy2019right}. We train the model on MNLI, and choose MNLI-mismatch and HANS as the ID and OOD test set.

For the fact verification task,
we use FEVER\footnote{
\url{https://github.com/TalSchuster/FeverSymmetric}} as the training dataset provided by \cite{thorne2018fever}. We train the model on FEVER, and evaluate the model performance on ID test set and FEVER Symmetric \cite{schuster2019towards} (version 1 and 2) as our OOD test set.

For the paraphrase identification task,
we perform the evaluation using Quora Question Pairs (QQP) as ID dataset and PAWS~\cite{zhang-etal-2019-paws} as OOD dataset which consists of two types of data including \emph{duplicate} and \emph{non-duplicate}.

\subsection*{Evaluation Metrics}
Following previous works, we measured the accuracy score on the in-distribution and out-of-distribution test sets to compare the results of different models.

\subsection{Implementation Details}
Following previous debiasing methods, we apply our debiasing method on the BERT-base \cite{devlin-etal-2019-bert}. The hyperparameters of BERT are consistent with previous research papers. The learning rate is 2e-5 for MNLI dataset and 1e-5 for FEVER and QQP, the batch size is 32 and the optimizer is AdamW with a weight decay of 0.01. Note that previous methods \cite{sanh2020learning,xiong2021uncertainty} have shown high variance in experiment results under different settings, we evaluate the performance of our model by four random seeds and report the averaged result. 
We use the \texttt{[CLS]} vector as sentence embedding for all three methods. 
The model is trained in an NVIDIA GeForce RTX 3090 GPU. All models are trained five epochs, and checkpoints with top-2 performance are finally evaluated on the challenge test set.

\subsection{Experimental Results}
The extensive results of all the above-mentioned methods are summarized in table~\ref{tab:exp1}. 
Compared with other baseline methods, Kernel-Whitening significantly improves the model performance on challenge sets, and achieves state-of-the-art results on seven of the eight benchmarks. On the MNLI and FEVER datasets, our framework achieves the best performance with about 10 percentage points higher than the accuracy of BERT-base, which outperforms other debiasing methods. This proves that our framework has the best results and generalizability among these methods.

Moreover, our approach can effectively eliminate the dataset bias while mitigating the damage to generalizable features. The vast majority of debiasing methods improve the performance of out-of-distribution datasets by sacrificing the performance of in-distribution datasets, which means that current debiasing methods attempt to achieve a trade-off between ID performance and OOD performance. However, our approach achieves the best performance on OOD datasets for natural language inference and fact verification tasks with better results on ID datasets. For QQP dataset, our proposed approach also achieves decent generalization in PAWS without excessive performance degradation on ID datasets. 

In general, the normalization methods perform well on both in-distribution and out-of-distribution datasets for all tasks. All five models of three methods are end-to-end approaches and do not rely on any prior knowledge of the dataset. That is to say, they achieve better utility and scalability while providing more effective debiasing. For BERT-whitening and Kernel-Whitening, a larger hidden dimension indicates better performance on OOD datasets, and Kernel-whitening performs better when parameters are constant to BERT-whitening, which strongly supports our analysis of normalization methods. 
BERT-flow shows an acceptable performance on OOD datasets, but is inferior to the whitening-based approach. We argue that flow model requires more samples as reference, and the original hyperparameters are not capable for the additional network layers.

%% file: outline/Discussion.tex
\section{Analysis and Discussion}

In this section, we construct supplementary experiments to further analyze the effectiveness of normalization methods, especially our Kernel-Whitening framework.

\subsection{Effect of Latent Dimension $L$}
\begin{figure}[t]
\centerline{\includegraphics[width=0.45 \textwidth]{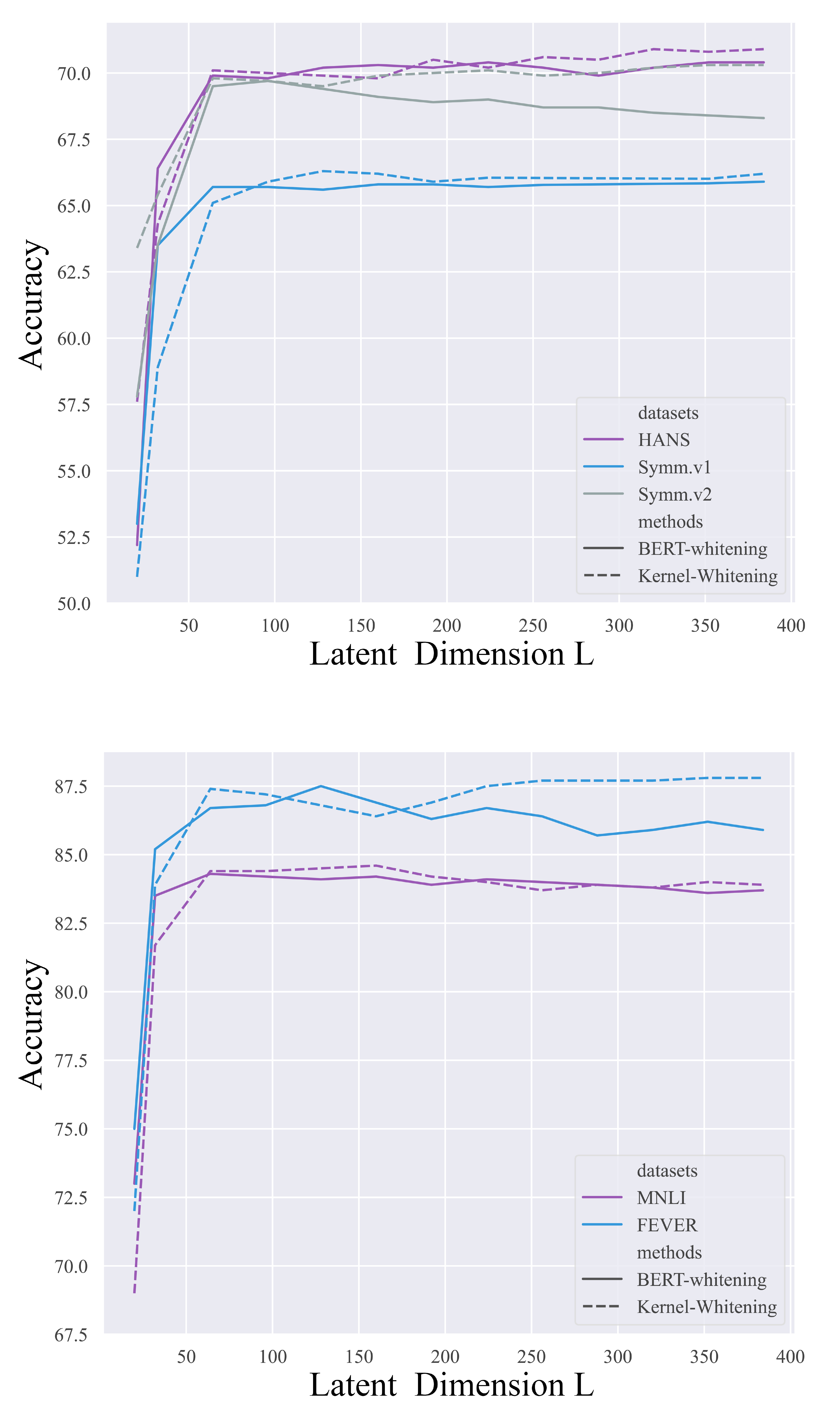}}
\caption{Effect of different dimensionality $L$ with whitening methods on each aforementioned tasks. The $x$ axis is the latent dimension of sentence embeddings. The two images are model performance on out-of-distribution and in-distribution test sets, respectively. }
\label{fig:dim}
\vspace{-1em}
\end{figure}

The dimensionality of reconstructed features is a key feature. The reduction of vector size brings smaller memory occupation and faster inference downstream layers, while the missing information may impair the ability of the model.
To further illustrate the effect of low-rank kernel approximation, we conduct a sensitivity analysis of latent dimension $L$. Figure \ref{fig:dim} shows the variation curve of performance change for two whitening-based methods. For both in-distribution and out-of-distribution tasks. A latent dimension of double the batch size provides promising performance. As the dimensionality rises, Kernel-Whitening maintains a stable debiasing effect, while BERT-whitening fluctuates on FEVER and Symm. v2 dataset. We argue that this phenomenon is due to that high-dimensional features are more prone to nonlinear correlations, where Kernel-Whitening is designed to show better results. Moreover, Kernel-Whitening always performs better when the dimensionality is greater than 300, which illustrates the stability and generality of our method.

\subsection{Independence Study}
\begin{figure}[t]
\centerline{\includegraphics[width=0.42 \textwidth]{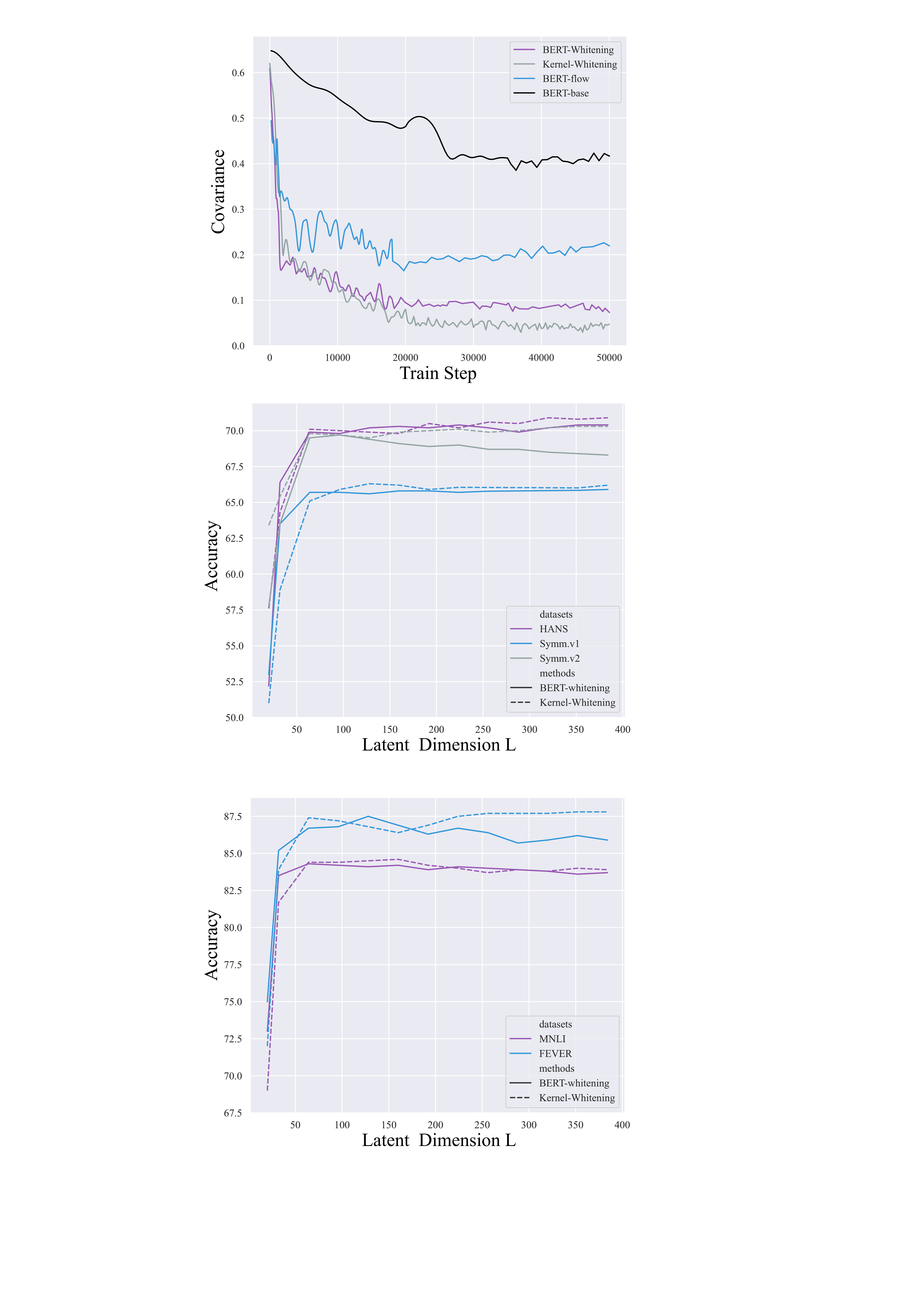}}
\caption{ The independence study on covariance between reconstructed features. The hidden dimension of BERT-whitening and Kernel-Whitening are set to 64. All models are trained 50000 steps with batch 32.}
\label{fig:cov}
\vspace{-1em}
\end{figure}

In Section \ref{sec:2} we analyse how isotropic data distribution leads to better generalization. To check whether normalization methods remove the dependencies between features, we conduct experiments on the covariance between features during training process. As shown in Figure \ref{fig:cov}, All three normalization methods exhibit a suppression effect on feature correlation, while our method achieves the optimal performance at the end of training. As the iterations increase, the covariance first decreases rapidly and converges to a low point. All method's performance fluctuations around certain steps, we believe such fluctuations are related to biased samples in the data.

Overall, Kernel-Whitening largely remits dependencies between features, and such independence effectively contributes to the generalization ability of deep network models.

\subsection{Time Consumption}
Besides outstanding debiasing performance, we compute the time consumption with baseline methods to further demonstrate the strength of Kernel-Whitening. We train each model equally on an NVIDIA RTX 2080Ti GPU with the same batch size. We compare three normalization methods with the best baseline work, MoCaD \cite{xiong2021uncertainty}, which trains a bias model to produce model calibrating. To give a horizontal comparison between different datasets, we set the time consumption of fine-tune 100 as a baseline. 
As shown in Table \ref{tab:time}, the time consumption of Kernel-Whitening is nearly the same as fine-tuning and costs 6 times less extra time than MoCad. Although BERT-whitening only uses a linear transformation to obtain reconstruction representations, our method is still faster. Because our method performs SVD decomposition on a matrix of $L*L$ while BERT-whitening handles the same operations on a matrix of $L*N$, where $L$ is the hidden dimension and $N$ is the output dimension of BERT ($e.g.$, 768).

\begin{table}[htbp] 
\begin{tabular}{l|c|c}
\toprule
\multicolumn{1}{c|}{\textbf{Method}} & \textbf{MNLI} & \textbf{FEVER}  \\
\midrule
{BERT-base} & 100  & 100\\
\midrule
{MoCad} & 264  & 227\\
{BERT-flow} & 198  & 190 \\
{BERT-whitening} & 146    & 139 \\
{\textbf{Kernel-Whitening}} & \textbf{138}  & \textbf{134}\\
\bottomrule
\end{tabular}
\centering
  \caption{Time consumption (percentages) of training one epoch on the whole dataset. Whitening-based methods cost much less time than previous works.}\label{tab:time}
\end{table}

%% file: outline/related.tex
\section{Related Work}

\subsection{Dataset Bias}

Recent observations \cite{mccoy2019right,naik2018stress} show that, natural language understanding models tend to over-rely on specific shallow heuristics, resulting in inadequate generalization capability in out-of-distribution (OOD) settings \cite{schuster2019towards}. \citet{sinha2021unnatural, pham2021out} have reported the insensitivity to word-order permutations among transformer-based models. The original and out-of-order examples elicit the same classification label When permuted randomly, which contradicts the conventional understanding of semantics. Such phenomena are studied as dataset bias problems.

Existing methods train additional models to identify biased training data \cite{clark2019don,utama2020mind,schuster2019towards} or use the above bias model to calibrate the classification results of test data \cite{utama2020towards,sanh2020learning,xiong2021uncertainty}. The so-called bias model refers to classifiers who use only a portion of input data for prediction, $e.g.$, hypothesis-only model in NLI task which only predict from specific linguistic phenomena in hypothesis sentences such as negation. These methods are not end-to-end and face difficulty in fully identifying all bias patterns.

Recently, another line of works notice the connection between dataset bias and feature distribution, and try to tackle the dataset bias problem by identifying features with better generalizability. \citet{dou2022decorrelate} use an loss function based on information bottleneck (IB) to focus the model on task-relevant features, and \citet{wu2022less} similarly achieve such feature filtering by mapping sentence embedding into a specific low-dimension subspace.

\subsection{Unsupervised Semantic of Sentence Embedding}
Previous works suggest that the word representations of pre-train language model are not isotropic \cite{gao2018representation,ethayarajh2019contextual}, leading model to poorly capture the underlying semantic of sentences \cite{li2020sentence}. 
Such anisotropic causes the difficulty of using sentence embedding directly through simple similarity metrics. \citet{gao2018representation} propose word embedding matrix regularization methods to mitigate the degeneration problem. Recently, researchers attempt to transform BERT sentence embedding into an isotropic Gaussian distribution through normalizing flow \cite{li2020sentence} or whitening methods \cite{su2021whitening}. As supervised learning also suffers from uneven data distribution of train sets, we are the first to normalize the data distribution on supervised training to eliminate dataset bias problem.

%% file: outline/conclusion.tex
\section{Conclusion}

In this work, we propose a novel framework, Kernel-Whitening, to tackle the spurious correlation from a feature perspective. We analyze how to introduce isotropic sentence embedding for eliminating dataset bias and propose a promising and computationally kernel estimation, to obtain an approximation of disentangled sentence embedding. Experiments on various datasets demonstrate that Kernel-Whitening achieves better performance on both ID and OOD datasets than comparative works. This implies that a shallow model, or prior knowledge of dataset bias, is not must for the improvement of generalization.


%% file: outline/limitation.tex
\section{Limitations}
In this section, we discuss the potential limitations of our work. The analysis of model effects in this paper is focusing on commonly used benchmarks for natural language understanding debiasing works, and they may carry confounding factors that affect the performance of our model.  Therefore, it is worth further exploring the performance of our model on more tasks, $e.g.$, the WikiGenderBias dataset for gender bias on relation extraction task. In addition, this presented work is inspired by unsupervised semantic learning methods, such as BERT-whitening, and it will be better to test the performance of our approach on unsupervised tasks. We leave these two problems to further work.